\documentclass[review, 3p]{elsarticle}
\usepackage{amssymb}
\usepackage{enumerate}
\usepackage{graphicx}
\usepackage{multirow}
\usepackage{amsmath}

\usepackage{algorithm}
\usepackage{algorithmic}
\renewcommand{\algorithmicrequire}{ \textbf{Input:}}     
\renewcommand{\algorithmicensure}{ \textbf{Output:}}    

\usepackage{array}
\usepackage{setspace}
\usepackage{tabularx}
\usepackage[colorlinks,linkcolor=blue,anchorcolor=blue,citecolor=blue]{hyperref}
\usepackage{enumitem}
\usepackage{xcolor}
\usepackage{colortbl}
\usepackage[caption=false,font=normalsize,labelfont=sf,textfont=sf]{subfig}
\usepackage{rotating}
\usepackage{wrapfig} 
\usepackage{flushend} 

\journal{ }

\begin{document}

\begin{frontmatter}

\title{Linear Gaussian Bounding Box Representation and Ring-Shaped Rotated Convolution for Oriented Object Detection}

\author[label1,label2]{Zhen Zhou}
\ead{zhouzhen2021@ia.ac.cn}

\author[label1,label2]{Yunkai Ma}

\author[label1]{Junfeng Fan}

\author[label1,label2]{Zhaoyang Liu}

\author[label1,label2]{Fengshui Jing\corref{cor1}}

\author[label1,label2]{Min Tan}

\address[label1]{State Key Laboratory of Multimodal Artificial Intelligence Systems, Institute of Automation, Chinese Academy of Sciences, No.95 Zhongguancun East Road, Beijing, 100190, China.}
\address[label2]{School of Artificial Intelligence, University of Chinese Academy of Sciences, No.19(A) Yuquan Road, Beijing, 100049, China.}



\begin{abstract}
In oriented object detection, current representations of oriented bounding boxes (OBBs) often suffer from boundary discontinuity problem. Methods of designing continuous regression losses do not essentially solve this problem. Although Gaussian bounding box (GBB) representation avoids this problem, directly regressing GBB is susceptible to numerical instability. We propose linear GBB (LGBB), a novel OBB representation. By linearly transforming the elements of GBB, LGBB avoids the boundary discontinuity problem and has high numerical stability. In addition, existing convolution-based rotation-sensitive feature extraction methods only have local receptive fields, resulting in slow feature aggregation. We propose ring-shaped rotated convolution (RRC), which adaptively rotates feature maps to arbitrary orientations to extract rotation-sensitive features under a ring-shaped receptive field, rapidly aggregating features and contextual information. Experimental results demonstrate that LGBB and RRC achieve state-of-the-art performance. Furthermore, integrating LGBB and RRC into various models effectively improves detection accuracy.

\end{abstract}




\end{frontmatter}

\vspace{\fill}\newpage



\section{Introduction}
\label{Introduction}

Detection of oriented objects is vital in a wide range of visual recognition scenarios, such as aerial image detection \cite{Oriented_RCNN}, text recognition \cite{TextBoxes++}, retail merchandise detection \cite{DRN}, etc. Different from horizontal bounding boxes (HBBs) \cite{Cascaded_Feature_Fusion, mSODANet}, oriented bounding boxes (OBBs) can provide more accurate object location information. Although great progress has been made in OBB detection in recent years \cite{RoI_Transformer, Rotated_cascade_R_CNN, R3Det, SCRDet}, accurate prediction of orientation information remains a challenge. To better extract orientation-related information, current methods primarily focus on the design of reasonable parameterized representations of OBBs \cite{KLD_PAMI, RSDet, Oriented_reppoints, CSL} and the extraction of rotation-sensitive features \cite{S2ANet, RRD, ReDet, ARC}.

\begin{figure*}[t]
  \centering
  \includegraphics[scale=0.670]{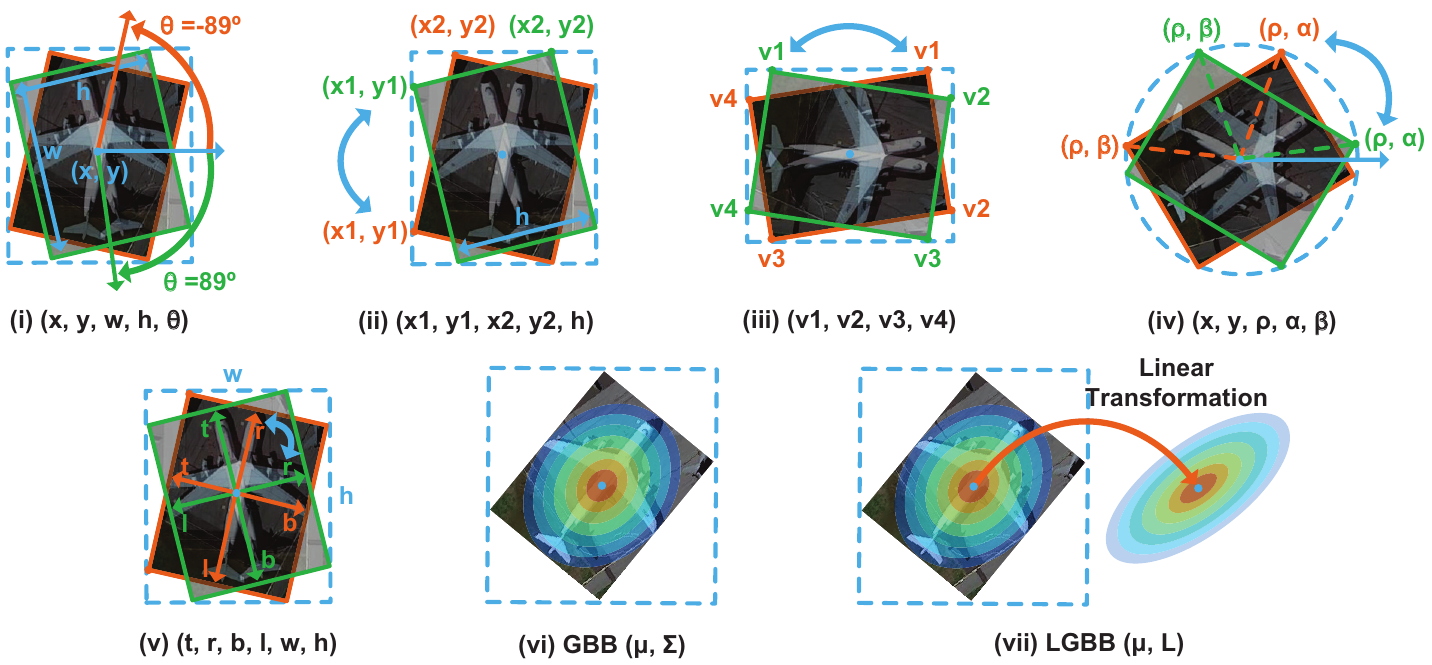}
  \caption{Seven different OBB representations. Different from some OBB representations suffer from the boundary discontinuity problem ((\romannumeral1)-(\romannumeral5)), GBB (\romannumeral6) and LGBB (\romannumeral7), which are modeled based on Gaussian distributions, do not have such problem. Compared with GBB, LGBB achieves high numerical stability by linearly transforming the elements of GBB.}
  \label{fig:OBB_Representation}
\end{figure*}

Since OBB only adds orientation information to HBB, regression on the combination of a HBB term and an orientation term is first commonly used \cite{DRN, RoI_Transformer, R3Det, SCRDet, RSDet}. However, due to the periodicity of angles, such methods suffer from the boundary discontinuity problem \cite{KLD_PAMI}, i.e., loss value sharply increases at boundary positions. As shown in Fig. \ref{fig:OBB_Representation} (\romannumeral1), at the boundary position, the predicted OBB (orange box) and ground truth (green box) are geometrically close, but their angle values are significantly different, making the regression loss of the angle term very large. By exploiting the geometric properties of OBB, many alternative representations of OBB have been proposed, such as two vertices and height of an OBB \cite{TextBoxes++, R2CNN} (Fig. \ref{fig:OBB_Representation} (\romannumeral2)), four vertices of a quadrilateral \cite{RRD, RIDet, Gliding_Vertex, ICN} (Fig. \ref{fig:OBB_Representation} (\romannumeral3)), two polar angles and polar radius of an OBB \cite{P_RSDet, PolarDet} (Fig. \ref{fig:OBB_Representation} (\romannumeral4)), and box boundary-aware vectors of an OBB \cite{BBAVectors} (Fig. \ref{fig:OBB_Representation} (\romannumeral5)). However, these OBB representations also suffer from the boundary discontinuity problem. Their boundary discontinuity cases are shown in Fig. \ref{fig:OBB_Representation} (\romannumeral2)-(\romannumeral5). 

To address the boundary discontinuity problem of OBB representations, many studies focus on designing continuous regression losses to alleviate or avoid such OBB representation problem. RSDet \cite{RSDet} adds a modulated term to the loss in boundary cases. SCRDet \cite{SCRDet} introduces IoU-smooth L1 loss to eliminate the sudden increase in loss at the boundary position. GWD \cite{GWD}, KLD \cite{KLD_PAMI} and SGKLD \cite{SGKLD} model OBB as Gaussian distribution and use the distance between two Gaussian distributions as the loss. The Gaussian distribution representation is continuous at boundary positions. However, these methods do not essentially solve the boundary discontinuity problem. Although the regression losses are boundary-continuous, the OBB representations that constitute the losses still suffer from this problem. For example, in Fig. \ref{fig:OBB_Representation} (\romannumeral1), the Gaussian distance-based loss between the predicted box and the ground truth is small, which guides the predicted box to rotate counterclockwise to the ground truth. However, due to the periodicity and value range (e.g., $-90^\circ$  to $90^\circ$) of the angle, the predicted box can only be rotated clockwise to regress to the ground truth (counterclockwise rotation will exceed the defined angle range), so the angle loss is still large. Other OBB representations that suffer from the boundary discontinuity problem have similar situations. Hence, to solve the boundary discontinuity problem, the ideal way is to find a continuous OBB representation.

\begin{figure}[t]
  \centering
  \includegraphics[scale=0.290]{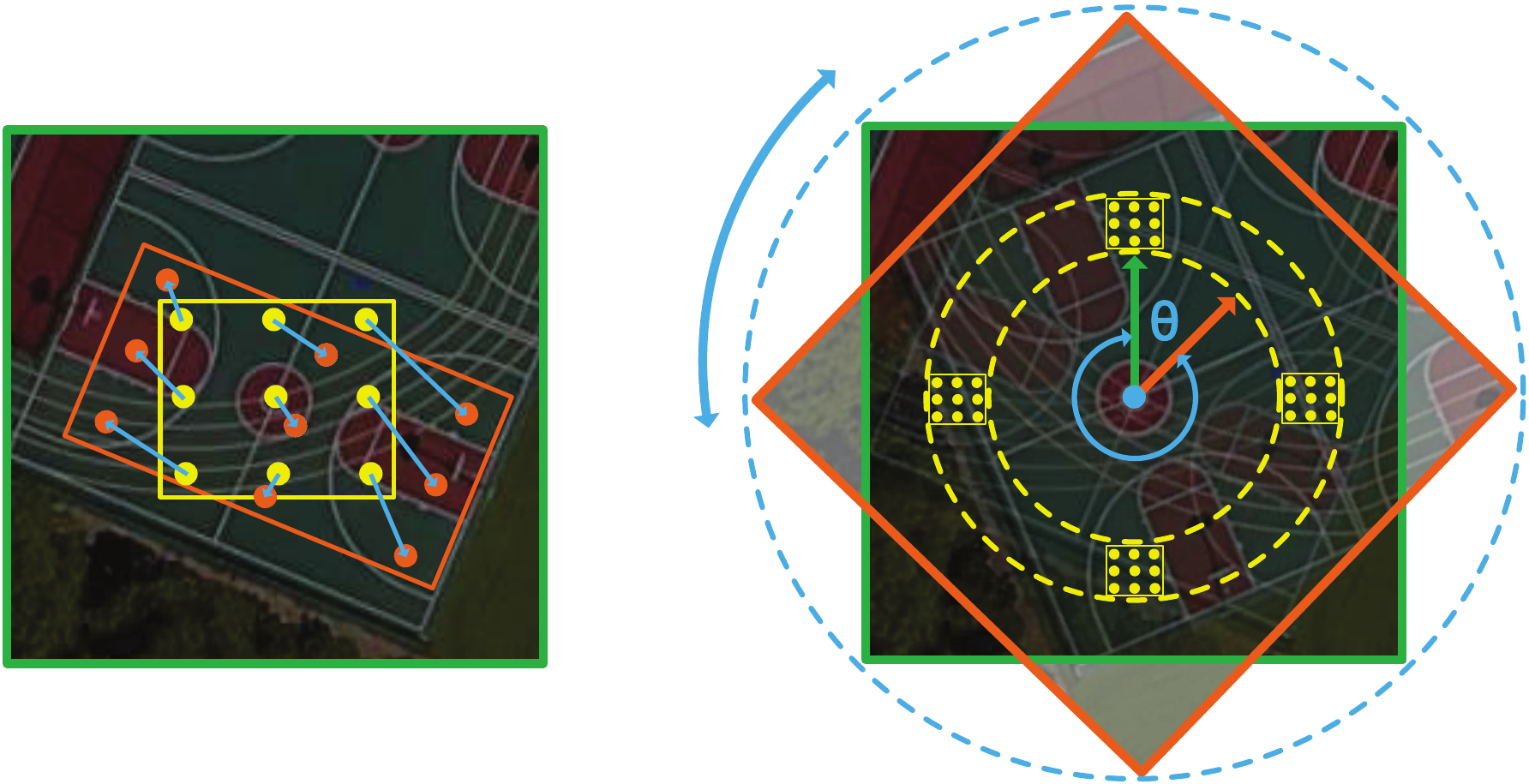}
  \caption{Two different types of rotation-sensitive feature extraction methods. Left: By improving standard convolution kernels (e.g., rotating convolution kernels) or adjusting original feature maps (e.g., deformable convolution \cite{DCN}). Right: By adaptively rotating feature maps. During the feature map rotation process, the receptive field of convolution is ring-shaped.}
  \label{fig:Orientation_Sensitive_Feature}
\end{figure}

On the other hand, extracting rotation-sensitive features helps models focus more on orientation-related information, thereby better adapting to complex orientation changes and predicting orientation information. Since standard convolution cannot extract features in various orientations well, variations of the standard convolution mode have been widely studied. Current methods mainly extract rotation-sensitive features from two aspects, i.e., improving standard convolution kernels and adjusting original feature maps, as shown in Fig. \ref{fig:Orientation_Sensitive_Feature}. The convolution kernels are rotated to different orientations \cite{S2ANet, RRD, ReDet, ARC} to extract information from multiple orientations on the feature maps. Inspired by deformable convolutions \cite{DCN}, some studies \cite{DRN, S2ANet, ICN, R3O, Deformable_Faster_RCNN} improve deformable convolutions to make them more suitable for oriented object detection. Different from rotating the convolution kernels, the methods based on deformable convolutions first generate the offsets of each kernel and then adjust feature maps, so that the convolution kernels can extract features in arbitrary orientations. Although these rotation-sensitive feature extraction methods can effectively enhance the ability to detect oriented objects, they are limited by the fact that convolutions can only extract local receptive field information. Convolution-based methods are slow in aggregating rotation-sensitive features and contextual information, which limits the detection performance of these convolution-based methods.

In this paper, we propose linear Gaussian bounding box (LGBB), a novel OBB representation which does not have the boundary discontinuity problem of OBB representations. As mentioned in \cite{KLD_PAMI, GBB}, Gaussian distribution representation (see Fig. \ref{fig:OBB_Representation} (\romannumeral6)) is continuous at boundary positions. Different from \cite{KLD_PAMI}, we choose to regress Gaussian bounding box (GBB) like \cite{GBB}. However, when directly regressing GBB and using the distance between Gaussian distributions (such as KLD \cite{KLD_PAMI} and ProbIoU \cite{GBB}) as the regression loss for GBB, it is susceptible to numerical instability. For example, when regressing GBBs of very small or large objects, it will produce very large or small gradients. LGBB does not directly regress GBB and does not use Gaussian distance-based regression losses to avoid numerical instability problem. To achieve high numerical stability and avoid the boundary discontinuity problem, LGBB linearly transforms the elements of GBB, as shown in Fig. \ref{fig:OBB_Representation} (\romannumeral7). Furthermore, to ensure the positive definiteness of the covariance matrix in GBB, a positive definite constraint term is added to the final regression loss. 

To overcome the shortcoming of slow aggregation of rotation-sensitive features and contextual information in convolution-based methods, we propose ring-shaped rotated convolution (RRC), which extracts rotation-sensitive features under a ring-shaped receptive field, rapidly aggregating rotation-sensitive features and contextual information. Unlike rotating convolutional kernels \cite{S2ANet, RRD, ReDet, ARC}, RRC extracts rotation-sensitive features by adaptively rotating feature maps. While directly rotating feature maps brings relatively high computational complexity, it enables the model to aggregate global contextual information within a ring-shaped region (see Fig. \ref{fig:Orientation_Sensitive_Feature}). Specifically, RRC divides feature maps into multiple rotation groups and adaptively rotates the feature maps within each rotation group to arbitrary orientations to extract rotation-sensitive features. Subsequently, the extracted rotation-sensitive features in multiple rotation groups will be combined for feature aggregation. During the feature map rotation process, the receptive field of convolution is ring-shaped. To reduce computational complexity and deeply extract more comprehensive rotation-sensitive features, rotation channel reduction and multi-scale group convolutions are applied. 

Experiments on DOTA-v1.5 \cite{DOTA} and HRSC2016 \cite{HRSC} datasets validate the effectiveness of the proposed LGBB and RRC. In summary, the main contributions of this paper are as follows: 
\begin{itemize}
  \item We propose linear Gaussian bounding box (LGBB), a novel OBB representation which does not have the boundary discontinuity problem and achieves high numerical stability. 
  \item We propose ring-shaped rotated convolution (RRC), which extracts rotation-sensitive features under a ring-shaped receptive field, rapidly aggregating rotation-sensitive features and contextual information. RRC can be applied to various models in a plug-and-play manner.
  \item The proposed LGBB and RRC achieve state-of-the-art (SOTA) performance. Furthermore, integrating LGBB and RRC into various models effectively improves detection accuracy.
\end{itemize}

The rest of this paper is organized as follows. Related works are discussed in Section \ref{Related Works}. Then, more details about our proposed methods are introduced in Section \ref{Methods}. Section \ref{Experiments} presents datasets, implementation details and experiments. Finally, we conclude in Section \ref{Conclusions}.

\section{Related Works}
\label{Related Works}

\subsection{Oriented Bounding Box Representation}

To represent an OBB, the simplest way is to add orientation term to a HBB. For example, SCRDet \cite{SCRDet}, R3Det \cite{R3Det} and RSDet \cite{RSDet} represented an OBB as ($x, y, w, h, \theta$), where ($x, y$), $w$, $h$ and $\theta$  represent the center point, width, height and orientation, respectively. However, such OBB representation suffers from the boundary discontinuity problem \cite{KLD_PAMI}. By exploiting the geometric properties of OBB, many alternative OBB representations have been proposed. TextBoxes++\cite{TextBoxes++} and R2CNN \cite{R2CNN} used the height and two vertices to represent an OBB. RIDet \cite{RIDet}, Gliding Vertex \cite{Gliding_Vertex}, RRD \cite{RRD} and ICN \cite{ICN} regressed four vertices of a quadrilateral to detect an OBB. P-RSDet \cite{P_RSDet} and PolarDet \cite{PolarDet} represented OBB in the polar coordinate system and detected the polar radius and two polar angles. BBAVectors \cite{BBAVectors} represented an OBB as box boundary-aware vectors. However, these OBB representations also have the boundary discontinuity problem. To address this problem, CSL \cite{CSL} and DCL \cite{DCL} converted the OBB regression task into a classification task. The OBB detection accuracy is influenced by the sampling interval. CFA \cite{CFA} and Oriented Reppoints \cite{Oriented_reppoints} used point set to represent OBB. Point set-based representations are sensitive to isolated points. Probability map-based methods \cite{Mask_OBB, CenterMap} treat the OBB detection task as a segmentation task. Mask OBB \cite{Mask_OBB} and CenterMap \cite{CenterMap} represented OBB as binary map and center probability map, respectively. However, probability map-based methods suffer from representation ambiguity problem. Pixels in the overlapping area are assigned multiple probability values of multiple objects simultaneously, making such representations ambiguous. GBB \cite{GBB} modeled OBB as Gaussian distribution. Although GBB avoids the boundary discontinuity problem, it is susceptible to numerical instability. To achieve high numerical stability, we propose LGBB by linearly transforming the elements of GBB. Furthermore, LGBB does not have the boundary discontinuity problem.

\subsection{Rotation-Sensitive Feature Extraction}

Since the standard convolution kernel cannot extract features in various orientations well, a natural idea is to improve the standard convolution kernel so that it can extract rotation-sensitive information in feature maps from multiple orientations. S2A-Net \cite{S2ANet} and RRD \cite{RRD} used active rotating filters (ARFs) \cite{ARF} to obtain rotation-sensitive features with explicitly encoded rotation information and further extract rotation-invariant features. Based on group equivariant convolutions \cite{GroupEC}, ReDet \cite{ReDet} proposed the rotation-equivariant backbone to extract rotation-equivariant features. ARC \cite{ARC} adaptively rotated convolution kernels to extract object features with varying orientations. Different from improving the standard convolution kernels, some other methods change the receptive field of the standard convolution kernel by adjusting original feature maps. ICN \cite{ICN} and Deformable Faster RCNN \cite{Deformable_Faster_RCNN} applied deformable convolutions directly to models to extract features in multiple orientations. S2A-Net \cite{S2ANet} introduced Alignment Convolution to enable the convolution kernel to extract rotation-sensitive information from arbitrary specific orientations. A similar idea was applied to the final prediction stage of the model \cite{R3O}. Similarly, based on deformable convolutions, DRN \cite{DRN} proposed rotation convolution layer (RCL) to extract rotation-sensitive information using multi-scale convolution kernels. However, the aforementioned rotation-sensitive feature extraction methods are limited by the fact that convolutions can only extract local receptive field information and are slow in aggregating rotation-sensitive features and contextual information. Our proposed RRC rotates original feature maps and extracts rotation-sensitive features under a ring-shaped receptive field, rapidly aggregating rotation-sensitive features and contextual information.

\section{Methods}
\label{Methods}

\begin{figure}[t]
  \centering
  \includegraphics[scale=0.320]{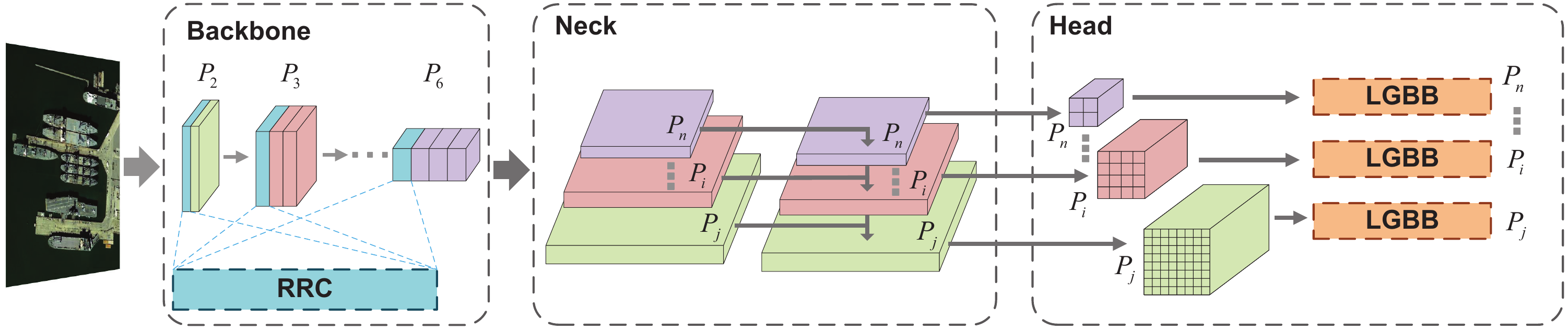}
  \caption{Layout of LGBB and RRC in oriented object detectors. For example, the regression target of the oriented object detector is LGBB, and RRC is applied to the first layer of each downsampling stage from $P_2$ to $P_6$ to extract rotation-sensitive features.}
  \label{fig:Overview}
\end{figure}

We first introduce LGBB representation in Section \ref{LGBB}, followed by the introduction of RRC in Section \ref{RRC}. The proposed RRC and LGBB can be conveniently applied to oriented object detectors, as shown in Fig. \ref{fig:Overview}. LGBB is used as the regression target of the model. RRC can be placed at any stage (e.g., backbone) of the model to extract rotation-sensitive features in a plug-and-play manner.

\subsection{Linear Gaussian Bounding Box}
\label{LGBB}

\begin{figure}[t]
  \centering
  \includegraphics[scale=0.650]{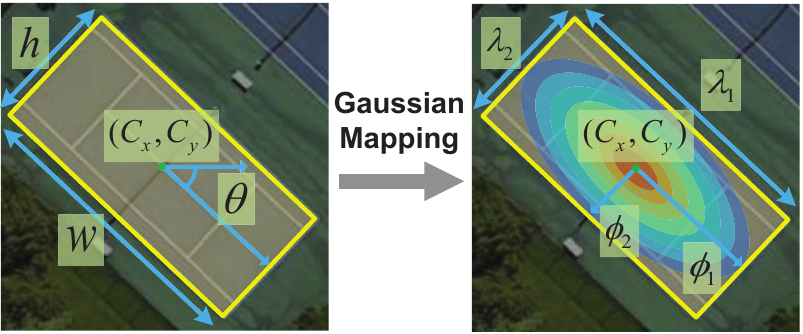}
  \caption{Mapping an OBB into a GBB.}
  \label{fig:LGBB_1}
\end{figure}

We first present how to model an OBB as a GBB. As shown in Fig. \ref{fig:LGBB_1}, given $(C_x, C_y)$, $w$, $h$ and $\theta$ as the center point, width, height and orientation of an OBB, respectively, a GBB ($G$) is parameterized as 
\begin{align}
  \mu &= (C_x, C_y)^T,  \\
  \Sigma 
  &=\begin{bmatrix}
    g_{1} & g_{2}\\
    g_{2} & g_{3}
  \end{bmatrix}
  = \begin{bmatrix}
    \phi_{1} & \phi_{2}
  \end{bmatrix}
  \begin{bmatrix}
    \lambda_{1} & \\
      & \lambda_{2}
  \end{bmatrix}
  \begin{bmatrix}
    \phi_{1} & \phi_{2}
  \end{bmatrix}^{T}
  =
  \begin{bmatrix}
    \lambda_{1} \cos^{2}{\theta} + \lambda_{2} \sin^{2}{\theta}  & (\lambda_{2} - \lambda_{1}) \sin{\theta}\cos{\theta} \\
    (\lambda_{2} - \lambda_{1}) \sin{\theta}\cos{\theta} & \lambda_{1} \sin^{2}{\theta} + \lambda_{2} \cos^{2}{\theta}
  \end{bmatrix}, \\
  G &= 
    \begin{bmatrix} 
      \mu & \Sigma
    \end{bmatrix} 
    = \begin{bmatrix}
        C_x & g_{1} & g_{2} \\
        C_y & g_{2} & g_{3}
      \end{bmatrix},
\end{align}
where $\lambda_{1}=\frac{w^2}{4}$ and $\lambda_{2}=\frac{h^2}{4}$. $\Sigma$ is a symmetric positive definite covariance matrix. We refer to the long edge definition ($D_{le}$) \cite{RoI_Transformer, ReDet} and let $w$ represents the long side. The 2-D Gaussian distribution is $\mathcal{N}$($\mu$, $\Sigma$). A GBB presents five degrees of freedom. More properties of $\Sigma$ are found in \cite{KLD_PAMI}. The value ranges of elements in $\Sigma$ are
\begin{align}
  \begin{aligned}
    g_{1}\in (\lambda_{2},\lambda_{1}), ~~~~ g_{2}\in (-\frac{\lambda_{1}-\lambda_{2}}{2},\frac{\lambda_{1}-\lambda_{2}}{2}), ~~~~ g_{3}\in (\lambda_{2},\lambda_{1}).
  \end{aligned}
\end{align}

The advantage of representing OBB as GBB is that GBB does not have the boundary discontinuity problem \cite{KLD_PAMI}. At any position (including boundary positions), GBB is continuous. Similar to \cite{GBB}, we choose to regress GBB. In \cite{GBB}, the convolution-based model directly regresses each element of GBB and uses the distance between two Gaussian distributions as the loss to guide the model output to regress to the target GBB. Specifically, the ProbIoU-based loss, which is based on Hellinger distance, is used to measure the distance between GBBs. However, directly regressing GBB is susceptible to numerical instability. We define
\begin{align}
  \underbrace{L_g = \mathcal{L}_g(\Sigma) = \mathcal{L}_g(g_1, g_2, g_3)}_{\mathrm{term}~A:~ \frac{\partial L_g}{\partial g_i}, ~i=1,2,3} = \underbrace{(\mathcal{L}_g \circ \mathcal{H}_g)(\lambda_1, \lambda_2, \theta)}_{\mathrm{term}~B:~ \mathcal{H}_g(\lambda_1, \lambda_2, \theta)},~~ (g_1, g_2, g_3) = \mathcal{H}_g(\lambda_1, \lambda_2, \theta),
\end{align}
where $L_g$ is the regression loss with respect to $\Sigma$. Since $\mu$ is less affected by numerical instability, we mainly consider $\Sigma$ here. The term A is the gradient value of $L_g$ with respect to each term in $\Sigma$. The term B represents the transformation function between $\Sigma$ and $(\lambda_1, \lambda_2, \theta)$. The numerical instability mainly arises from these two terms. 

For term A: 
\begin{itemize}
  \item When the model, which uses Gaussian distance-based loss (e.g., KLD or ProbIoU), is backpropagated during training, the denominator of $\frac{\partial L_g}{\partial g_i}$ ($i=1,2,3$) consists of side length terms (i.e., $\lambda_{1}$ and $\lambda_{2}$). This produces very large or small gradients when regressing very small or large objects.
\end{itemize}

For term B: 
\begin{itemize}
  \item All elements in $\Sigma$ are coupled with the orientation ($\theta$), and frequent changes in $\theta$ lead to instability in the regression of $\Sigma$.
  \item To ensure the positive definiteness of $\Sigma$ (i.e., $g_1, g_3 > 0$ and det $\Sigma=g_1 g_3 - g_2^2 > 0$), additional terms with large numerical ranges (such as exponential term) are often introduced, resulting in numerical instability at some extreme positions.
\end{itemize}

Numerical instability caused by the aforementioned reasons limits the OBB detection performance. To achieve high numerical stability, LGBB linearly transforms the elements of GBB, which is described as 
\begin{align}
  L_G &= 
  \begin{bmatrix} 
    \mu & L
  \end{bmatrix} 
  = \begin{bmatrix}
      C_x & l_{1} & l_{2} \\
      C_y & l_{2} & l_{3}
    \end{bmatrix},
\end{align}
where 
\begin{align}
  \begin{bmatrix}
    l_1 \\
    l_2 \\
    l_3 
  \end{bmatrix}
  = L_T 
  \begin{bmatrix}
    g_1 \\
    g_2 \\
    g_3 
  \end{bmatrix}
  = \begin{bmatrix}
      \frac{1}{2} & 0 & \frac{1}{2} \\
      1 & 0 & 0 \\
      \frac{1}{2} & 1 & \frac{1}{2} 
    \end{bmatrix}
    \begin{bmatrix}
      g_1 \\
      g_2 \\
      g_3 
    \end{bmatrix}.
\end{align}
$L_T$ is linear transformation matrix and the value ranges of elements in $L$ are
\begin{align}
  \begin{aligned}
    l_{1}=\frac{\lambda_{1}+\lambda_{2}}{2}, ~~~~ l_{2}\in (\lambda_{2},\lambda_{1}), ~~~~ l_{3}\in (\lambda_{2},\lambda_{1}).
  \end{aligned}
\end{align}

After linear transformation, the regression on LGBB has high numerical stability. We also define
\begin{align}
  &\underbrace{L_l = \mathcal{L}_l(L) = \mathcal{L}_l(l_1, l_2, l_3)}_{\mathrm{term}~A:~ \frac{\partial L_l}{\partial l_i}, ~i=1,2,3} = \underbrace{(\mathcal{L}_l \circ \mathcal{H}_l)(\lambda_1, \lambda_2, \theta)}_{\mathrm{term}~B:~ \mathcal{H}_l(\lambda_1, \lambda_2, \theta)},~~ (l_1, l_2, l_3) = \mathcal{H}_l(\lambda_1, \lambda_2, \theta),
\end{align}
where $L_l$ is the regression loss with respect to $L$. The term A is the gradient value of $L_l$ with respect to each term in $L$. The term B represents the transformation function between $L$ and $(\lambda_1, \lambda_2, \theta)$. LGBB mainly enhances the numerical stability of these two terms during the regression process.

\begin{figure}[t]  
  \centering
  \subfloat[\centering]{\includegraphics[scale=0.800]{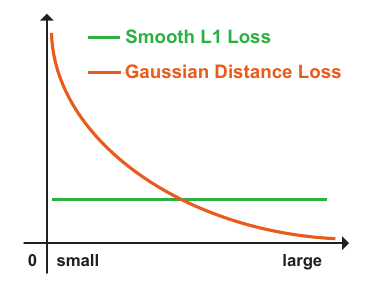}\label{fig:LGBB_2_a}}
  \subfloat[\centering]{\includegraphics[scale=0.900]{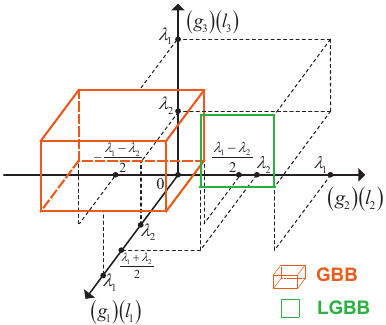}\label{fig:LGBB_2_b}}
  \caption{Comparisons between regression on GBB and regression on LGBB. (a) The relationship between the gradient values of the losses with respect to GBB and LGBB (ordinate) and the object sizes (abscissa). (b) The range of values for $\Sigma$ in GBB and $L$ in LGBB.}
\end{figure}

For term A:
\begin{itemize}
  \item Instead of using the distance between Gaussian distributions as the regression loss, we use Smooth L1 loss to avoid numerical instability caused by gradients, as shown in Fig. \ref{fig:LGBB_2_a}.
\end{itemize}

For term B:
\begin{itemize}
  \item Compared with GBB, $l_1$ in LGBB is decoupled from the orientation ($\theta$), which reduces numerical instability caused by frequent changes in $\theta$. 
  \item As shown in Fig. \ref{fig:LGBB_2_b}, $L$ is constrained within a smaller and more stable range, and directly regressing $L$ does not produce numerically unstable terms, such as exponential term. To ensure the positive definiteness of $\Sigma$ before the linear transformation, an additional positive definite constraint term without numerical instability is added to the regression loss.
  \item Compared with $g_2$ in $\Sigma$ which can be negative (see Fig. \ref{fig:LGBB_2_b}), the elements in $L$ are positive and constrained between $\lambda_{1}$ and $\lambda_{2}$, which is more conducive to the learning of anchor-based methods, such as YOLOv7 \cite{YOLOv7} and Oriented RCNN \cite{Oriented_RCNN}.
\end{itemize}

The linear transformation operation improves the shortcoming of directly regressing GBB from the perspective of numerical stability. Although the design of linear transformation matrix $L_T$ seems to be heuristic, it brings high numerical stability and effectively improves model learning efficiency, especially for anchor-based methods.

For the regression of LGBB, the weighted Smooth L1 loss is used as the loss for anchor-based methods, which is formulated as follows:
\begin{align}
  \Delta C_x &= \frac{C_x^p - C_x^a}{|w^a \ast \cos \theta^a| + |h^a \ast \sin \theta^a|} 
  - \frac{C_x^t - C_x^a}{|w^a \ast \cos \theta^a| + |h^a \ast \sin \theta^a|}, \\
  \Delta C_y &= \frac{C_y^p - C_y^a}{|w^a \ast \sin \theta^a| + |h^a \ast \cos \theta^a|} 
  - \frac{C_y^t - C_y^a}{|w^a \ast \sin \theta^a| + |h^a \ast \cos \theta^a|}, \\
  \Delta l_1 &= \log (l_1^p / l_1^a) - \log (l_1^t / l_1^a), \\
  \Delta l_2 &= \log (l_2^p / l_2^a) - \log (l_2^t / l_2^a), \\
  \Delta l_3 &= \log (l_3^p / l_3^a) - \log (l_3^t / l_3^a), \\
  L_{\mathrm{reg}}^{s} &= L_1^{\mathrm{smooth}} (\Delta C_x, \Delta C_y, \Delta l_1, \Delta l_2, \Delta l_3),
\end{align}
where $(C_x^p, C_y^p, l_1^p, l_2^p, l_2^p)$ and $(C_x^t, C_y^t, l_1^t, l_2^t, l_2^t)$ denote the predicted LGBB and the corresponding ground truth, respectively. $(C_x^a, C_y^a, l_1^a, l_2^a, l_2^a)$ represents the anchor assigned to the ground truth. $(w^a, h^a, \theta^a)$ is derived from $(l_1^a, l_2^a, l_2^a)$. $\Delta C_x$ and $\Delta C_y$ are scaled using the sloping edges of the OBB.

To ensure the positive definiteness of $\Sigma$ derived from the predicted LGBB, we introduce a positive definite constraint term which forces the model to regress towards the target in the right direction to the final regression loss ($L_{\mathrm{reg}}$), which is given as 
\begin{align}
  L_{\mathrm{reg}}&=\gamma_{1} L_{\mathrm{reg}}^{s} -\gamma_{2} \underbrace{\min\{l_{2}(2l_1-l_2)-(l_3-l_1)^2,0\}}_{\mathrm{positive} ~ \mathrm{definite} ~ \mathrm{constraint} ~ \mathrm{term}},
\end{align}
where $\gamma_{1}$ and $\gamma_{2}$ are the weights to trade off the two terms. The expression of the positive definite constraint term is derived from $g_1, g_3 > 0$ and det $\Sigma=g_1 g_3 - g_2^2 > 0$. Unfortunately, the isotropic Gaussian distribution cannot be oriented when an OBB is square. SGKLD \cite{SGKLD} models an OBB as a super-Gaussian distribution to avoid the case where the distribution is isotropic. However, learning the super-Gaussian distribution is difficult, and deriving the analytical solution of OBB from the super-Gaussian distribution is complex. Here, we propose a simple yet effective training trick to address this problem. When the OBB is close to a square, the long side of the target OBB is extended by 2\% during training. The prediction error for the original object is within 1\%, so the error effect can be neglected under the existing performance metrics. Therefore, target distributions are anisotropic.

The model directly predicts LGBB in the forward inference stage and maps it back to ($C_x, C_y, w, h, \theta$) in the post-processing stage.

\subsection{Ring-Shaped Rotated Convolution}
\label{RRC}

\begin{algorithm}{} 
	\renewcommand{\algorithmicrequire}{\textbf{Input:}}
	\renewcommand{\algorithmicensure}{\textbf{Output:}}
	\caption{RRC}
	\label{algorithm:AngleConv}
	\begin{algorithmic}[1]

      \REQUIRE $\mathcal{F}_I$
      
      \STATE $\mathcal{F}_H$ $\leftarrow$ Reduce rotation channel

      \STATE Generate $\varPhi$ using AGM

      \FOR{$F_h$ $\subset$ $F_H$}
          \STATE $F_R$ $\supset$ $F_{r}$ $\leftarrow$ Rotate $F_h$ according to $\varPhi$ and pad with 0
      \ENDFOR

      \STATE $\mathcal{F}_P$ $\leftarrow$ Fuse $\mathrm{PE}(X_H)$ + $\mathrm{PE}(Y_H)$ into $\mathcal{F}_R$ by PEM

      \STATE $\mathcal{F}_E$ $\leftarrow$ Extract rotation-sensitive features from $\mathcal{F}_P$ using RFEM

      \STATE $\mathcal{F}_H^{\ast}$ $\leftarrow$ Map $\mathcal{F}_E$ back according to ${\varPhi^{-1}}$

      \STATE $\mathcal{F}_I^{\ast}$ $\leftarrow$ Restore rotation channel

      \STATE $\mathcal{F}_O = \mathcal{F}_I + \mathcal{F}_I^{\ast}$

      \ENSURE $\mathcal{F}_O$

	\end{algorithmic}  
\end{algorithm}

\begin{figure*}[t]
    \centering
    \includegraphics[scale=0.680]{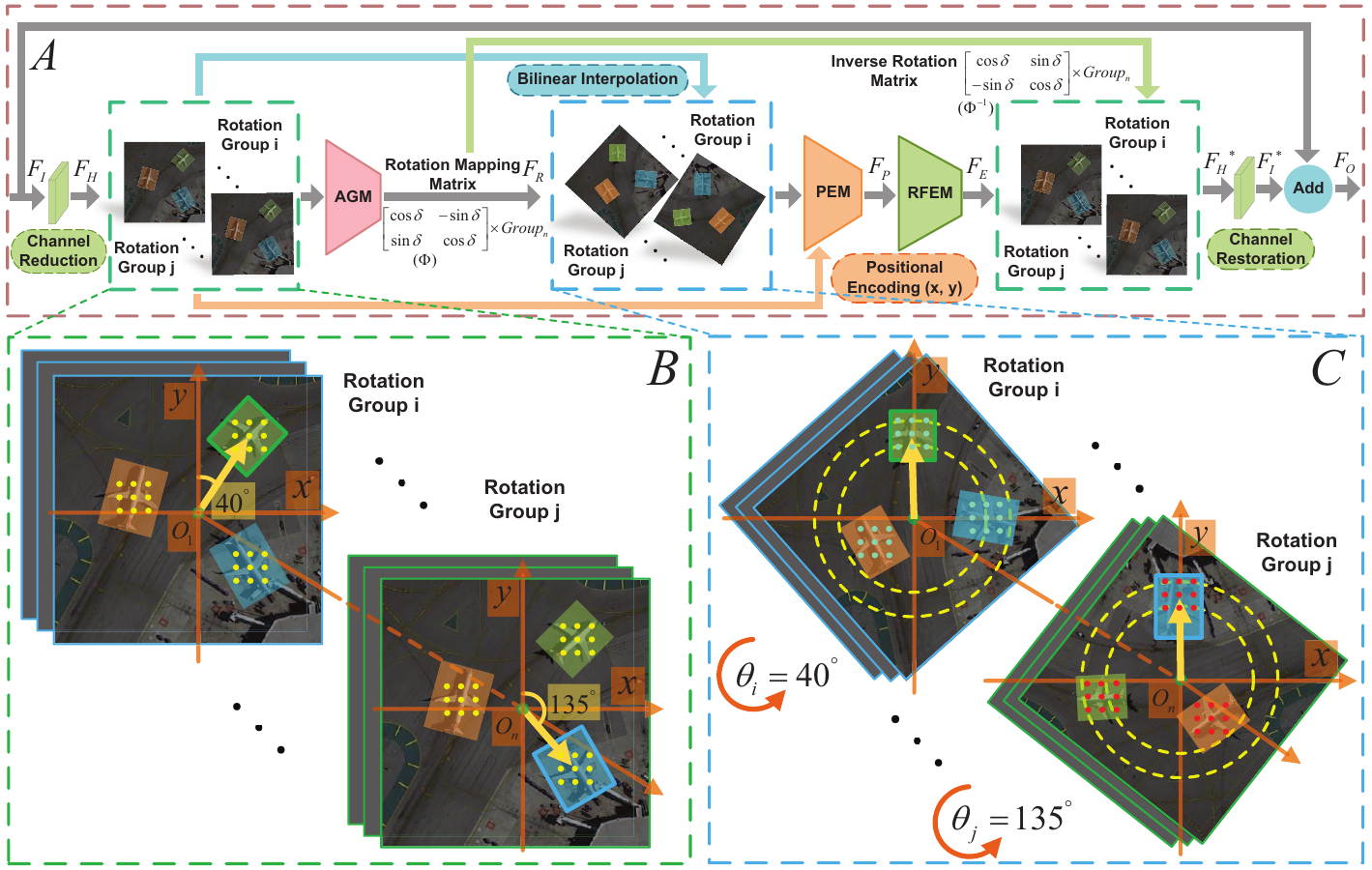} 
    \caption{Architecture of the proposed RRC. The panel A presents the overall process of RRC. Panels B and C show more details of feature maps before and after rotation, respectively. For clarity of expression, original images rather than intermediate feature maps are used to demonstrate the RRC implementation process.}
    \label{fig:RRC}
\end{figure*}
  
\begin{figure*}[t]
\centering
\includegraphics[scale=0.190]{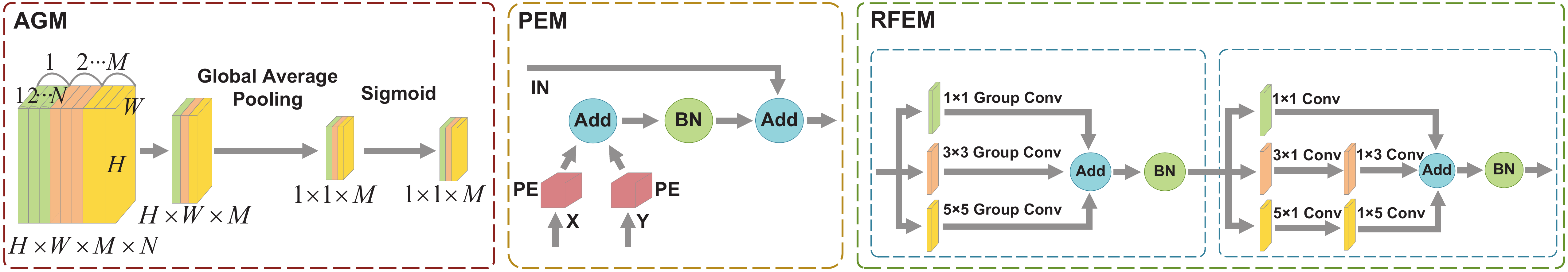}
\caption{The intermediate modules of RRC. Left: AGM is responsible for adaptively generating corresponding rotation angle for each rotation group. Middle: PEM is used to fuse the positional encoding information of the feature maps before rotation into the rotated feature maps. Right: RFEM is responsible for extracting rotation-sensitive features from the rotated feature maps and aggregating the extracted rotation-sensitive features and contextual information under a ring-shaped receptive field.}
\label{fig:RRC_module}
\end{figure*}

Differing from existing methods \cite{S2ANet, RRD, ReDet, ARC} that can only extract rotation-sensitive features under a local receptive field by rotating convolution kernels, RRC achieves a ring-shaped receptive field by adaptively rotating feature maps to arbitrary orientations and performing rotation-sensitive feature extraction and feature aggregation on the rotated feature maps. Fig. \ref{fig:RRC} demonstrates the structure of RRC. The pseudocode of RRC is described in Algorithm \ref{algorithm:AngleConv}. The main processes of RRC are as follows.

\subsubsection{Rotation Channel Reduction and Feature Map Grouping}
Due to the relatively high computational complexity introduced by rotating feature maps, RRC first conducts channel reduction on the input feature maps $\mathcal{F}_I \in \mathbb{R}^{H \times W \times C}$, where $H$, $W$ and $C$ represent the height, width and channel of $\mathcal{F}_I$, respectively. A $1 \times 1$ convolution is applied to obtain the feature maps $\mathcal{F}_H \in \mathbb{R}^{H \times W \times K}$ after channel reduction. Next, $\mathcal{F}_H$ is divided into $M$ rotation groups by channel for subsequent rotation-sensitive feature extraction. There are $N$ channels in each rotation group ($K=M \times N$). 

\subsubsection{Adaptive Rotation of Feature Maps}
To extract rotation-sensitive features in arbitrary orientations, the rotation angle is first adaptively generated for each rotation group by using an angle generation module (AGM). As depicted in Fig. \ref{fig:RRC_module}, the process of AGM is given as 
\begin{align}
  \mathcal{F}_A &= 2 \pi * \sigma(\mathrm{GAP}(\mathrm{Conv}_{3 \times 3} (\mathcal{F}_H))), \\
  \varPhi &=  
  \begin{bmatrix}
    \cos \mathcal{F}_a & -\sin \mathcal{F}_a \\
    \sin \mathcal{F}_a &  \cos \mathcal{F}_a 
  \end{bmatrix}, ~ \mathcal{F}_a \in \mathcal{F}_A,
\end{align}
where global average pooling (GAP) is applied to aggregate feature map information within each group, as designed in \cite{GAP}. $\mathrm{Conv}_{3 \times 3}$ denotes a $3 \times 3$ convolution. The Sigmoid function ($\sigma$) is used for rotation angle range constraint. Then, these rotation angles ($\mathcal{F}_A$) are transformed into rotation mapping matrices ($\varPhi$).

\begin{figure}[t]
  \centering
  \includegraphics[scale=0.640]{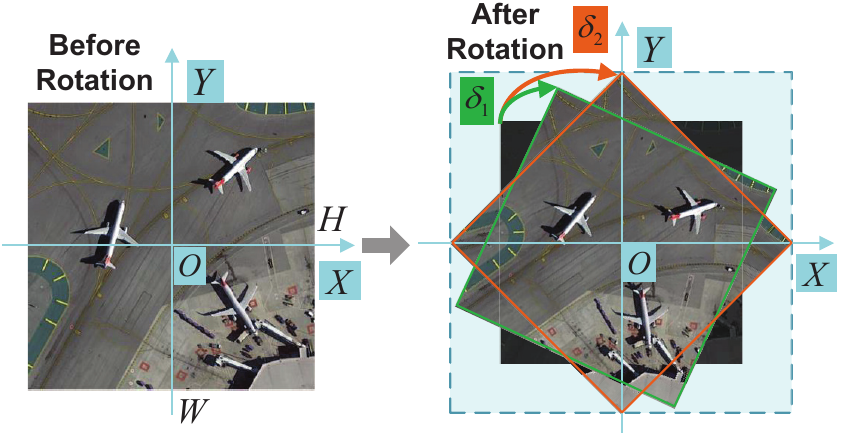}
  \caption{Illustration of the feature maps which are rotated to different orientations according to the rotation angles generated by AGM. The area inside the blue dashed box denotes the range of the large feature maps $\mathcal{F}_R$. All the rotated feature maps are contained in this area.}
  \label{fig:RRC_rotation}
\end{figure}

According to the rotation mapping matrices $\varPhi$, the feature maps of different rotation groups are rotated counterclockwise around their geometric centers. Feature maps within the same rotation group are rotated by the same angle. However, due to the inconsistent geometric shapes of the feature maps in different rotation groups after rotation, RRC uniformly constrains all feature maps and maps them to larger feature maps ($\mathcal{F}_R \in \mathbb{R}^{\sqrt{H^2+W^2} \times \sqrt{H^2+W^2} \times K}$). As shown in Fig. \ref{fig:RRC_rotation}, $\mathcal{F}_R$ consists of the following two parts: 
\begin{enumerate}[label=(\arabic*)]
  \item For elements rotated from $\mathcal{F}_H$, bilinear interpolation is used to fill them.
  \item Other elements are padded with a default value, e.g., 0.
\end{enumerate}

Furthermore, to enhance the information integrity of the feature maps before and after rotation, a positional encoding module (PEM) is introduced to fuse the position information of $\mathcal{F}_H$ to $\mathcal{F}_R$. PEM obtains the positional encodings of $\mathcal{F}_H$ using a shared convolution and then fuses them into $\mathcal{F}_R$, which is described as
\begin{align}
  \mathcal{F}_P = \mathrm{BN}(\mathrm{PE}(X_H) + \mathrm{PE}(Y_H)) + \mathcal{F}_R,
\end{align}
where $X_H$ and $Y_H$ represent the coordinates of $\mathcal{F}_H$, and $\mathcal{F}_P$ represents the fused feature maps. $\mathrm{PE}$ uses the $1 \times 1$ convolution. $\mathrm{BN}$ represents the combination of batch normalization layer and activation layer. Subsequent rotation-sensitive feature extraction and aggregation are performed on $\mathcal{F}_P$.

\subsubsection{Rotation-Sensitive Feature Extraction}
To deeply extract more comprehensive rotation-sensitive features under a ring-shaped receptive field, RRC introduces a rotation-sensitive feature extraction module (RFEM). Unlike ARC \cite{ARC}, which uses rotated convolution kernels to extract rotation-sensitive features, RFEM extracts rotation-sensitive features on the rotated feature maps. As shown in Fig. \ref{fig:RRC_module}, RFEM first uses multi-scale group convolutions ($\mathrm{GC}$) to extract rotation-sensitive features from the rotated feature maps $\mathcal{F}_P$, which is given as
\begin{align}
  \mathcal{F}_G = \mathrm{BN}(\mathrm{GC}_{1 \times 1}(\mathcal{F}_P) + \mathrm{GC}_{3 \times 3}(\mathcal{F}_P) + \mathrm{GC}_{5 \times 5}(\mathcal{F}_P)),
\end{align}
where $\mathcal{F}_G$ represents the extracted feature maps.

Multi-scale convolutions deeply mine more comprehensive features from multiple receptive fields. Group convolutions extract rotation-sensitive features within each rotation group separately and further reduce computational complexity and parameters. The number of groups is the same as the number of rotation groups. RRC adaptively generates continuous rotation angles based on the orientation information of input feature maps, extracting rotation-sensitive features more efficiently.    

\subsubsection{Feature Aggregation under a Ring-Shaped Receptive Field}
Subsequently, RFEM further aggregates the rotation-sensitive features extracted from each separate rotation group under a ring-shaped receptive field. As described in panel C of Fig. \ref{fig:RRC}, the trajectory of each position in feature maps during the rotation is a circle. If a convolution kernel with the size of $k \times k$ is used to extract features, the receptive field of a certain position is also $k \times k$. Therefore, during the rotation process, the receptive field of each position is a ring with a width of $k$. For example, if the yellow ring is regarded as the receptive field, the width of the ring is $k$. In other words, if a convolution is used to extract features across all rotation groups of the feature maps $\mathcal{F}_P$ (note that the rotation angle of each rotation group is different), this convolution is equivalent to having a ring-shaped receptive field. 

RFEM uses multi-scale convolutions across all rotation groups to aggregate rotation-sensitive features and contextual information, as depicted in Fig. \ref{fig:RRC_module}. This is described as
\begin{align}
  \mathcal{F}_E &= \mathrm{BN}(\mathrm{Conv}_{1 \times 1}(\mathcal{F}_G) + \mathrm{Conv}_{1 \times 3}(\mathrm{Conv}_{3 \times 1}(\mathcal{F}_G)) 
                +  \mathrm{Conv}_{1 \times 5}(\mathrm{Conv}_{5 \times 1}(\mathcal{F}_G))),
\end{align}
where $\mathcal{F}_E$ represents the extracted feature maps. To reduce computational complexity and parameters, RFEM decomposes a $k \times k$ convolution into a pair of $k \times 1$ and $1 \times k$ convolutions, as designed in \cite{k_1_1_k}. RFEM adaptively aggregates information at different positions among all rotation groups under a ring-shaped receptive field. For example, before rotation, the objects within the green and blue boxes are relatively far apart (see panel B). After rotation, the two objects are relatively close to each other (see panel C), so RRC can aggregate the rotation-sensitive features of the two objects at one time. The ring-shaped receptive field greatly enhances the ability to rapidly aggregate rotation-sensitive features and contextual information, thereby improving model learning efficiency.

\subsubsection{Inverse Mapping and Rotation Channel Restoration}
After rotation-sensitive feature extraction and aggregation, to eliminate the extra padded areas in $\mathcal{F}_R$ and preserve the same structure as $\mathcal{F}_H$, $\mathcal{F}_E$ needs to be mapped back into the feature maps before rotation pixel by pixel according to the inverse directions of the rotation mapping matrices $\varPhi^{-1}$. Finally, a $1 \times 1$ convolution is applied to the mapped feature maps ($\mathcal{F}_H^{\ast}$) to restore channels. The restored feature maps ($\mathcal{F}_I^{\ast}$) are then combined with the input feature maps $\mathcal{F}_I$ to generate outputs ($\mathcal{F}_O$) by using a skip connection \cite{ResNet}.

\section{Experiments}
\label{Experiments}

Section \ref{Datasets} introduces two datasets, DOTA-v1.5 \cite{DOTA} and HRSC2016 \cite{HRSC}, which are commonly used for oriented object detection tasks. Subsequently, some implementation details related to model parameters, training and inference settings are introduced in Section \ref{Implementation Details}. Then, a series of ablation experiments are conducted on the testing set of DOTA-v1.5 dataset to evaluate the effectiveness of the proposed LGBB and RRC (Section \ref{Ablation Studies}). Finally, the performance of our proposed methods is further verified by comparative experiments with some SOTA methods on these two datasets (Section \ref{Comparison_SOTA}). 

\subsection{Datasets}
\label{Datasets}

DOTA dataset is a large scale dataset for oriented object detection in aerial images. It contains 2806 large aerial images with sizes ranging from 800$\times$800 to 4000$\times$4000 and 402,089 instances among 16 common categories. Compared with DOTA-v1.0, DOTA-v1.5 contains more extremely small instances which are less than 10 pixels. The training set has 1411 images while the validation set contains 458 images. The testing set consists of 937 images. In our experiments, both the training set and validation set are employed for training, and the testing set without annotations is used for evaluation. The original images are cropped into 1024$\times$1024 patches with a stride of 500. At the training stage, the images are randomly flipped and rotated to avoid overfitting. For fair comparisons with other methods, the original images are resized at three scales (0.5, 1.0 and 1.5) for multi-scale training and testing.

HRSC2016 dataset is a high resolution ship detection dataset, which contains 1061 images ranging from 300$\times$300 to 1500$\times$900. We employ the training set (436 images) and validation set (181 images) for training and the testing set (444 images) for evaluation. All images are resized to 800$\times$800 without changing the aspect ratio. The images are randomly rotated and flipped for data augmentation during training.

\subsection{Implementation Details} 
\label{Implementation Details}

Our experiments are mainly conducted on the powerful horizontal object detector YOLOv7 and mmrotate \cite{mmrotate} framework. YOLOv7 is improved to detect oriented objects by adding an oriented head, which predicts orientation-related information. Specifically, the proposed RRC is applied to the first layer of each downsampling stage from $P_2$ to $P_6$ to extract rotation-sensitive features (see Fig. \ref{fig:Overview}). The regression target is the LGBB representation of oriented objects. All structures are built on YOLOv7-W6. For anchor settings, considering the variability of the geometric shapes of oriented objects, the aspect ratios of preset anchors are set to $2 \colon 1$ and $5 \colon 1$. The rotation angles are set to $0^{\circ}$, $45^{\circ}$, $90^{\circ}$ and $135^{\circ}$. 

For more details, the stochastic gradient descent (SGD) optimizer is applied for training and the initial learning rate is set to 0.001 with the warming up for 500 iterations. The momentum is set to 0.9, and the weight decay is $10^{-4}$. We train the model for 40 epochs for the DOTA-v1.5 dataset and 80 epochs for the HRSC2016 dataset. In the loss, the weights of the confidence term, box term and classification term are set to 0.4, 0.5 and 0.1, respectively. The hyperparameters of focal loss \cite{focal_loss} are set to $\alpha=0.25$ and $\gamma=2.0$. The code related to YOLOv7 is available at \href{https://github.com/zhen6618/RotaYolo}{https://github.com/zhen6618/RotaYolo}.

For the experimental settings on the mmrotate framework, we follow the same training settings as the base models, such as Oriented RCNN. Like YOLOv7, the proposed LGBB serves as the regression target. RRC is applied to the first layer of each downsampling stage from $P_2$ to $P_6$ to extract rotation-sensitive features.

All models are first pretrained on MS COCO dataset for 280 epochs and then pretrained on DOTA-v1.5 HBB Task for 20 epochs. Finally, they are refined on the DOTA-v1.5 OBB Task and HRSC2016 dataset. The batch size is set to 8 (2 images per GPU) for training and 1 for testing. All experiments are conducted on a server with 4 RTX 2080Ti GPUs for training and deployed on a single RTX 3080Ti GPU for inference.

\subsection{Ablation Studies}
\label{Ablation Studies}

To evaluate the effectiveness of the components of the proposed LGBB and RRC, a series of ablation experiments are conducted on the DOTA-v1.5 dataset. Multi-scale training is not used in this part. YOLO and RX-FPN stand for using YOLOv7 and ResNetX \cite{ResNet} with a feature pyramid structure as backbones, respectively.

\begin{table}[t]
  \footnotesize
  \begin{center}
  \begin{tabular}{m{7.0em}<{\centering} m{2.5em}<{\centering} m{2.5em}<{\centering} m{2.5em}<{\centering} m{2.5em}<{\centering} m{2.5em}<{\centering} m{2.5em}<{\centering} m{2.5em}<{\centering} m{2.5em}<{\centering}}
    \hline
    Backbone & \multicolumn{2}{c}{YOLO} & \multicolumn{2}{c}{R50-FPN} & \multicolumn{2}{c}{R101-FPN} & \multicolumn{2}{c}{R152-FPN}\\ 
    \hline
    Representation & $\Sigma$ & $L$ & $\Sigma$ & $L$ & $\Sigma$ & $L$ & $\Sigma$ & $L$ \\
    \hline
    mAP(\%) & 64.33 & \textbf{66.12} & 63.59 & \textbf{65.20} & 63.72 & \textbf{65.34} & 63.86 & \textbf{65.51} \\
    \hline
  \end{tabular}
  \end{center}
  \caption{Comparison of regression results on GBB and LGBB.}
  \label{table:Linear_Transformation}
\end{table}

\subsubsection{Linear Transformation}
Instead of directly regressing GBB, LGBB linearly transforms the elements of GBB. To evaluate the effectiveness of the proposed linear transformation, GBB and LGBB regression losses are computed using ProbIoU-based loss and Smooth L1 loss, respectively. We directly regress ($g_1$, $g_2$, $g_3$). As described in Table \ref{table:Linear_Transformation}, the performance of regression on $L$ is better than the performance of regression on $\Sigma$ on all backbones. By using linear transformation, the detection performance is significantly improved. 

This is mainly attributed to two benefits of the linear transformation: (1) Compared with GBB, where some elements have negative values and a wide value range, each element in LGBB has a positive value and a more stable value range, which is more beneficial for anchor-based model learning. (2) LGBB has only two terms coupled with orientations, resulting in less interference from frequently changing orientations during model training. (3) Compared with Gaussian distance-based loss, Smooth L1 loss avoids numerical instability caused by gradients. These advantages make LGBB more numerically stable than GBB, resulting in better detection performance on all backbones.

\subsubsection{Positive Definite Constraint} 
To ensure the positive definiteness of $\Sigma$ derived from the predicted LGBB, the positive definite constraint term is introduced to the final regression loss. We compare the results with and without the positive definite constraint in Table \ref{table:positive_definite}. The hyperparameters used to trade off the Smooth L1 term and positive definite constraint term are set to $\gamma_{1}$ = 0.8 and $\gamma_{2}$ = 0.2. In addition, we also compare the constraint method used in \cite{GBB}, which regresses additional exponential functions and unconstrained parameters. The regression loss is defined using the Smooth L1 loss. 

Using a positive definite constraint term for LGBB regression effectively improves detection accuracy, which achieves 66.38\% mAP when YOLO serves as the backbone. The positive definite constraint term encourages models to learn in correct regression directions. Compared with the constraint method used in \cite{GBB}, LGBB regression with the positive definite constraint has higher numerical stability.

\begin{table}[t]
  \footnotesize
  \begin{center}
  \begin{tabular}{m{6.0em}<{\centering} m{17em}<{\centering} m{4.5em}<{\centering}}
    \hline
    Backbone & Positive Definite Constraint Method & mAP(\%) \\ 
    \hline
    \multirow{3}{*}{YOLO} & Constraint Method in \cite{GBB} & 64.37\\
    & Without Positive Definite Constraint & 66.12 \\
    & With Positive Definite Constraint & \textbf{66.38} \\
    \hline
    \multirow{3}{*}{R50-FPN} & Constraint Method in \cite{GBB} & 64.09\\
    & Without Positive Definite Constraint & 65.20 \\
    & With Positive Definite Constraint & \textbf{65.51} \\
    \hline
    \multirow{3}{*}{R101-FPN} &Constraint Method in \cite{GBB} & 64.38\\
    & Without Positive Definite Constraint & 65.34 \\
    & With Positive Definite Constraint & \textbf{65.62} \\
    \hline
    \multirow{3}{*}{R152-FPN} & Constraint Method in \cite{GBB} & 64.26\\
    & Without Positive Definite Constraint & 65.51 \\
    & With Positive Definite Constraint & \textbf{65.69} \\
    \hline
  \end{tabular}
  \end{center}

  \caption{Performance of different methods for ensuring the positive definite property of $\Sigma$.}
  \label{table:positive_definite}
\end{table}

\subsubsection{Ring-Shaped Rotation Convolution} 
While directly rotating feature maps brings relatively high computational complexity, RRC enables the model to capture global rotation-sensitive information within a ring-shaped region. During RRC forward inference, its computational complexity mainly depends on the number of channels of the input feature maps. Therefore, in RRC, channel reduction is first performed on input feature maps. In addition, RRC uses the group convolution to extract rotation-sensitive features separately in each rotation group, which also reduces computational complexity and parameters.

\begin{table}[t]
  \footnotesize
  \begin{center}
  \begin{tabular}{m{6.0em}<{\centering} m{3.5em}<{\centering} m{3.5em}<{\centering} m{4.5em}<{\centering} m{4.5em}<{\centering} m{4.5em}<{\centering}}
    \hline
    Backbone & K & M & mAP(\%) & GFLOPs & Params\\ 
    \hline
    \multirow{11}{*}{YOLO} & - & - & 66.38 & 93.74 & 36.62M \\
    \cline{2-6}
    & \multirow{5}{*}{32} & 1 & 67.88 & 109.94 & 38.12M \\
    && 2 & 67.95 & 106.83 & 38.03M \\
    && 4 & 68.13 & 105.32 & 37.99M \\
    && 8 & \textbf{68.18} & \textbf{104.63} & \textbf{37.97M} \\
    && 16 & 67.97 & 104.44 & 37.97M \\
    \cline{2-6}
    & \multirow{5}{*}{64} & 1 & 67.95 & 150.48 & 39.39M \\
    && 2 & 68.05 & 137.99 & 39.03M \\
    && 4 & 68.15 & 131.83 & 38.86M \\
    && 8 & \textbf{68.22} & \textbf{128.89} & \textbf{38.78M} \\
    && 16 & 68.04 & 127.73 & 38.75M \\
    \hline
  \end{tabular}
  \end{center}

  \caption{Comparison results of RRC with different parameter settings and the standard convolution. ``-'' represents the standard convolution.}
  \label{table:RRC_complexity}
\end{table}

\begin{figure}[t]
  \centering
  \includegraphics[scale=0.25]{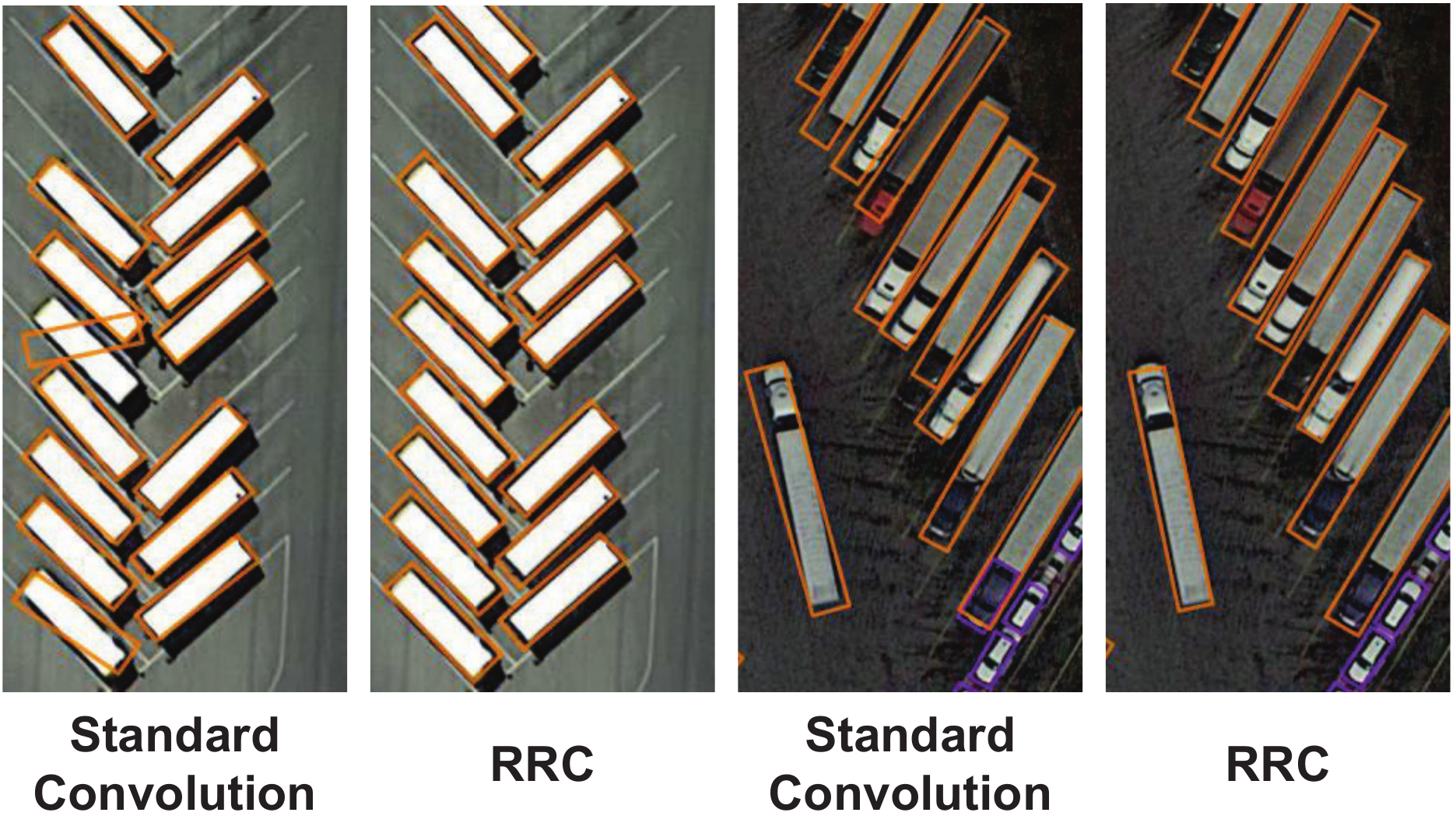}
  \caption{Comparison results between RRC and standard convolution for rotation-sensitive feature extraction. The yellow and purple bounding boxes indicate the predictions of large vehicles and small vehicles, respectively.}
  \label{fig:RRC_vis}
\end{figure}

\begin{table}[t]
  \footnotesize
  \begin{center}
  \begin{tabular}{m{2.2em}<{\centering} m{2.2em}<{\centering} m{2.2em}<{\centering} m{2.2em}<{\centering} m{2.2em}<{\centering} m{2.2em}<{\centering} m{2.2em}<{\centering} m{8.0em}<{\centering}}
    \hline
    \multicolumn{7}{c}{Component} & \multicolumn{1}{c}{Metric}\\ 
    \hline
    CR & AGM & PEM & MS & GC & SC & AGG & mAP(\%) \\
    \hline
     &  &  &  &  &  &  & 66.38 \\
    \checkmark &  &  &  &  &  &  & 66.69 \textcolor{blue}{($\uparrow$ 0.31)} \\
    \checkmark & \checkmark &  &  &  &  &  & 67.02 \textcolor{blue}{($\uparrow$ 0.64)} \\
    \checkmark & \checkmark & \checkmark &  &  &  &  & 67.17 \textcolor{blue}{($\uparrow$ 0.79)} \\
    \checkmark & \checkmark & \checkmark & \checkmark &  &  &  & 67.38 \textcolor{blue}{($\uparrow$ 1.00)} \\
    \checkmark & \checkmark & \checkmark & \checkmark & \checkmark &  &  & 67.54 \textcolor{blue}{($\uparrow$ 1.16)} \\
    \checkmark & \checkmark & \checkmark & \checkmark & \checkmark & \checkmark &  & 67.76 \textcolor{blue}{($\uparrow$ 1.38)} \\
    \checkmark & \checkmark & \checkmark & \checkmark & \checkmark & \checkmark & \checkmark & \textbf{68.22 \textcolor{blue}{($\uparrow$ 1.84)}} \\
    \hline
  \end{tabular}
  \end{center}

  \caption{Performance of each component of RRC.}
  \label{table:RRC_component}
\end{table}

Experimental results are shown in Table \ref{table:RRC_complexity}, and LGBB is the regression target. When the number of reduced channels ($K$) and the number of rotation groups ($M$) are set to 32 and 8, respectively, the accuracy reaches 68.18\% mAP, which is 1.80\% mAP higher than standard convolution. RRC significantly improves detection performance, and the increased computational complexity and parameters are acceptable. The comparison results of RRC and standard convolution are visualized in Fig. \ref{fig:RRC_vis}.

We also verify the effectiveness of each component of RRC in Table \ref{table:RRC_component}. YOLO serves as the backbone. Each component designed in RRC helps improve the performance of rotation-sensitive feature extraction. By using channel reduction (CR) and adaptive rotation of feature maps to extract information from arbitrary orientations, RRC significantly enhances the ability to extract rotation-sensitive features. PEM helps track the position information of feature maps before and after rotation. In RFEM, using multi-scale (MS) group convolutions (GC) for each rotation group to separately extract rotation-sensitive features improves the accuracy by 0.37\% mAP. The skip connection (SC) in the final step of RRC also effectively alleviates training instability and accelerates the training process. Most importantly, by aggregating (AGG) rotation-sensitive features and contextual information under a ring-shaped receptive field, the detection accuracy is significantly improved by 0.46\% mAP. Feature aggregation with a larger receptive field enables RRC to rapidly extract rotation-sensitive features, thereby speeding up model training and improving model learning efficiency.

\begin{table}[t]
  \footnotesize 
  \begin{center}
  \begin{tabular}{m{7.5em}<{\centering} m{5.0em}<{\centering} m{5.0em}<{\centering} m{5.0em}<{\centering} m{5.0em}<{\centering}}
    \hline
    Representation & YOLO & R50-FPN & R101-FPN & R152-FPN\\ 
    \hline
    ($x,y,w,h,\theta$)        & 64.77 & 63.91 & 64.00 & 63.84 \\
    ($x_1,y_1,x_2,y_2,h$)     & 64.34 & 63.16 & 63.66 & 63.59 \\
    ($v_1,v_2,v_3,v_4$)       & 64.12 & 63.84 & 64.04 & 63.67 \\
    ($x,y,\rho,\alpha,\beta$) & 64.38 & 63.37 & 63.50 & 64.07 \\
    ($t,r,b,l,w,h$)           & 64.56 & 63.62 & 63.74 & 64.18 \\
    GBB ($\mu, \Sigma$)       & 65.89 & 64.88 & 65.03 & 64.91 \\
    LGBB ($\mu, L$) & \textbf{66.38} & \textbf{65.51} & \textbf{65.62} & \textbf{65.69} \\
    \hline
  \end{tabular}
  \end{center}

  \caption{Performance comparison among different OBB representations.}
  \label{table:Comparison_OBB_Repre}
\end{table}

\subsection{Comparisons with the State-of-the-Art}
\label{Comparison_SOTA}

In this part, we compare the proposed LGBB and RRC with current SOTA methods on the DOTA-v1.5 and HRSC2016 datasets. More experimental details are presented in Section \ref{Datasets} and Section \ref{Implementation Details}.

\subsubsection{Oriented Bounding Box Representation} 
To verify the effectiveness of the proposed LGBB, we compare it with some current OBB representations (corresponding to the seven different OBB representations in Fig. \ref{fig:OBB_Representation}) on the DOTA-v1.5 dataset. RRC is not used in this experiment. Apart from GBB and LGBB, all other OBB representations use Smooth L1 loss as the regression loss. The implementation of regression on GBB refers to \cite{GBB}. The regression loss on LGBB consists of the Smooth L1 loss term for regressing the bounding box and the positive definite constraint term.   

As shown in Table \ref{table:Comparison_OBB_Repre}, experimental results demonstrate that LGBB achieves the best performance on all backbones. Compared with OBB representations without Gaussian distribution modeling, LGBB improves detection accuracy by at least 1\% mAP. This is primarily attributed to the fact that LGBB does not have the boundary discontinuity problem. Furthermore, compared with GBB, LGBB has higher numerical stability, which contributes to the efficient learning for anchor-based models.

\subsubsection{Rotation-Sensitive Feature Extraction}
To evaluate the effectiveness of the proposed RRC, we compare it with current other types of convolutions for rotation-sensitive feature extraction on the DOTA-v1.5 dataset. The regression target is the LGBB. The number of orientation channels in both Rotation-equivariant Convolution (ReConv) \cite{ReDet} and Oriented Response Convolution (ORConv) \cite{ARF} is set to 8. For fair comparison, except for RRC, all other convolutions use $1 \times 1$, $3 \times 3$ and $5 \times 5$ multi-scale convolution kernels to extract rotation-sensitive features. Like RRC, all convolutions are applied only to the first layer of each downsampling stage from $P_2$ to $P_6$. 

\begin{table}[t]
  \footnotesize
  \begin{center}
  \begin{tabular}{m{10.0em}<{\centering} m{5.0em}<{\centering} m{5.0em}<{\centering} m{5.0em}<{\centering}}
    \hline
    Convolution Type & R50-FPN & R101-FPN & R152-FPN \\ 
    \hline
    Standard Conv      & 65.51 & 65.62 & 65.69 \\
    Deformable Conv    & 66.07 & 66.21 & 66.24 \\
    ReConv             & 66.59 & 66.47 & 66.50 \\
    ORConv             & 66.34 & 66.51 & 66.43 \\
    ARC                & 66.85 & 66.72 & 67.08 \\
    RRC       & \textbf{67.21} & \textbf{67.04} & \textbf{67.36} \\
    \hline
  \end{tabular}
  \end{center}

  \caption{Performance comparison among different rotation-sensitive feature extraction methods.}
  \label{table:Comparison_Rotation_sensitive}
\end{table}

Experimental results in Table \ref{table:Comparison_Rotation_sensitive} demonstrate that RRC outperforms other convolutions in rotation-sensitive feature extraction. Compared with other types of convolutions, RRC rapidly extracts rotation-sensitive features and aggregates these features and contextual information under a ring-shaped receptive field. This enhances model learning efficiency and detection performance. Adding only a few layers of RRC to the model effectively improves the ability to extract rotation-sensitive features.

\begin{sidewaystable}
  \scriptsize

  \begin{center}
  \begin{tabular}{m{13.5em}<{\centering} m{6.3em}<{\centering} m{2.0em}<{\centering} m{2.0em}<{\centering} m{2.0em}<{\centering} m{2.0em}<{\centering} m{2.0em}<{\centering} m{2.0em}<{\centering} m{2.0em}<{\centering} m{2.0em}<{\centering} m{2.0em}<{\centering} m{2.0em}<{\centering} m{2.0em}<{\centering} m{2.0em}<{\centering} m{2.0em}<{\centering} m{2.0em}<{\centering} m{2.0em}<{\centering} m{2.0em}<{\centering} m{2.3em}<{\centering}}
    \hline
    Method & Backbone & PL & BD & BR & GTF & SV & LV & SH & TC & BC & ST & SBF & RA & HA & SP & HC & CC & mAP \\ 
    \hline
    RetinaNet-O\cite{focal_loss} & R50-FPN & 75.67 & 83.28 & 48.35 & 69.55 & 48.77 & 60.81 & 78.85 & 90.86 & 80.82 & 66.24 & 56.37 & 70.34 & 67.58 & 69.14 & 53.67 & 11.62 & 64.50 \\
    
    Faster RCNN-O\cite{DOTA} & R101-FPN & 76.79 & 82.23 & 50.37 & 66.21 & 57.13 & 75.03 & 85.69 & 89.99 & 81.26 & 71.81 & 53.97 & 70.74 & 67.01 & 72.35 & 66.51 & 16.47 & 67.72 \\
    
    Mask OBB\cite{Mask_OBB} & R101-FPN & 83.56 & 78.57 & 57.02 & 64.49 & 58.97 & 78.71 & 85.45 & 90.40 & 80.31 & 73.65 & 53.61 & 71.52 & 70.49 & 69.58 & 64.73 & 23.81 & 69.05  \\
    
    Gliding Vertex\cite{Gliding_Vertex} & R50-FPN & 88.13 & 83.68 & 54.21 & 76.87 & 64.34 & 70.49 & 85.63 & 90.15 & 77.66 & 73.86 & 64.25 & 74.14 & 71.28 & 67.09 & 63.17 & 29.25 & 70.89 \\
    
    CSL\cite{CSL} & R101-FPN & 88.72 & 83.25 & 55.26 & 76.32 & 63.87 & 70.54 & 79.99 & 90.41 & 84.01 & 73.83 & 72.43 & 73.94 & 72.41 & 67.70 & 70.56 & 37.50 & 72.55  \\
    
    R3Det\cite{R3Det} & R101-FPN & 88.20 & 83.02 & 53.15 & 68.73 & 67.52 & 81.59 & 88.91 & 90.56 & 81.65 & 73.61 & 70.22 & 71.83 & 69.34 & 73.88 & 74.48 & 39.15 & 73.49  \\
  
    SCRDet\cite{SCRDet} & R50-FPN & 88.24 & 87.97 & 57.62 & 74.94 & 62.34 & 81.51 & 82.54 & 90.77 & 83.96 & 74.82 & 71.01 & 67.47 & 73.93 & 71.11 & 75.46 & 40.90 & 74.04 \\
    
    RoI Transormer\cite{RoI_Transformer} & R101-FPN & 88.53 & 87.49 & \textcolor{blue}{61.86} & 81.20 & 67.75 & 79.69 & 89.93 & 90.55 & 80.12 & 73.12 & 68.52 & 74.25 & \textcolor{blue}{78.43} & 72.51 & 74.31 & 43.12 & 75.71 \\
    
    S2A-Net\cite{S2ANet} & R101-FPN & 88.62 & 87.56 & 58.72 & \textcolor{blue}{81.96} & 68.11 & 81.23 & \textcolor{blue}{90.24} & 90.67 & 80.86 & 75.42 & 71.98 & 73.13 & 77.75 & 72.86 & 72.38 & 43.58 & 75.94  \\
    \hline

    YOLO \cite{YOLOv7}-LR & YOLO & 88.74 & 87.01 & 61.52 & 79.65 & 68.30 & 80.98 & 89.74 & 89.97 & 81.12 & 75.82 & 72.36 & 75.25 & 77.48 & 73.62 & 73.24 & 50.26 & \textbf{76.57} \\
    \hline
  
    Oriented Reppoints \cite{Oriented_reppoints} & R101-FPN & 89.12 & 88.06 & 61.56 & 80.45 & 68.14 & 81.47 & 89.67 & 90.47 & 81.68 & 75.13 & 71.25 & 74.58 & 77.74 & 74.66 & 74.79 & 46.62 & 76.59 \\

    Oriented Reppoints \cite{Oriented_reppoints}-R & R101-FPN & 89.27 & 87.95 & 61.78 & 80.79 & \textcolor{blue}{68.74} & 81.59 & 89.71 & 90.66 & 82.49 & \textcolor{blue}{76.01} & 71.07 & 74.83 & 77.80 & 75.12 & 74.52 & 47.10 & \textbf{76.84} \\
    \hline
  
    ReDet \cite{ReDet} & R50-FPN & 88.51 & 86.45 & 61.23 & 81.20 & 67.60 & \textcolor{blue}{83.65} & 90.00 & 90.86 & 84.30 & 75.33 & 71.49 & 72.06 & 78.32 & 74.73 & 76.10 & 46.98 & 76.80 \\

    ReDet \cite{ReDet}-L & R50-FPN & 88.94 & 86.27 & 61.35 & 81.09 & 68.22 & 83.14 & 89.98 & \textcolor{blue}{90.88} & \textcolor{blue}{84.70} & 75.92 & 71.58 & 74.14 & 78.20 & \textcolor{blue}{75.18} & 77.04 & 46.73 & \textbf{77.09} \\
    \hline
    
    R3Det-KLD \cite{KLD_PAMI} & R50-FPN & 89.23 & 87.17 & 60.77 & 79.98 & 67.84 & 81.86 & 89.63 & 89.93 & 82.88 & 75.11 & 72.75 & 74.67 & 76.54 & 74.71 & 78.02 & 48.73 & 76.86 \\

    R3Det-KLD \cite{KLD_PAMI}-LR & R50-FPN & 89.31 & 87.54 & 60.52 & 80.17 & 67.99 & 81.63 & 89.86 & 90.24 & 82.87 & 75.23 & 72.54 & 74.65 & 76.95 & 75.02 & \textcolor{blue}{78.13} & 52.79 & \textbf{77.22} \\
    \hline
  
    Oriented RCNN \cite{Oriented_RCNN} & R101-FPN & 89.35 & 88.24 & 60.86 & 79.67 & 68.43 & 81.22 & 89.95 & 89.92 & 81.74 & 75.59 & \textcolor{blue}{73.10} & \textcolor{blue}{75.31} & 78.38 & 74.27 & 77.83 & 48.66 & 77.03 \\

    Oriented RCNN \cite{Oriented_RCNN}-LR & R101-FPN & \textcolor{blue}{89.37} & \textcolor{blue}{88.27} & 61.85 & 80.77 & 68.34 & 81.03 & 89.85 & 89.94 & 81.25 & 75.72 & 73.01 & 75.18 & 78.16 & 74.34 & 76.95 & \textcolor{blue}{56.18} & \textcolor{blue}{\textbf{77.51}}\\
    \hline
  
  \end{tabular}
  \end{center}

  \caption{Comparison results on the DOTA-v1.5 dataset. ``-L'' and ``-R'' represent LGBB and RRC, respectively. ``-LR'' stands for LGBB and RRC. The result with \textcolor{blue}{blue} color represents the best result in each column. The same applies below.}
  \label{table:DOTA_detection}
\end{sidewaystable}

\begin{figure*}[t]
  \centering
  \includegraphics[scale=0.520]{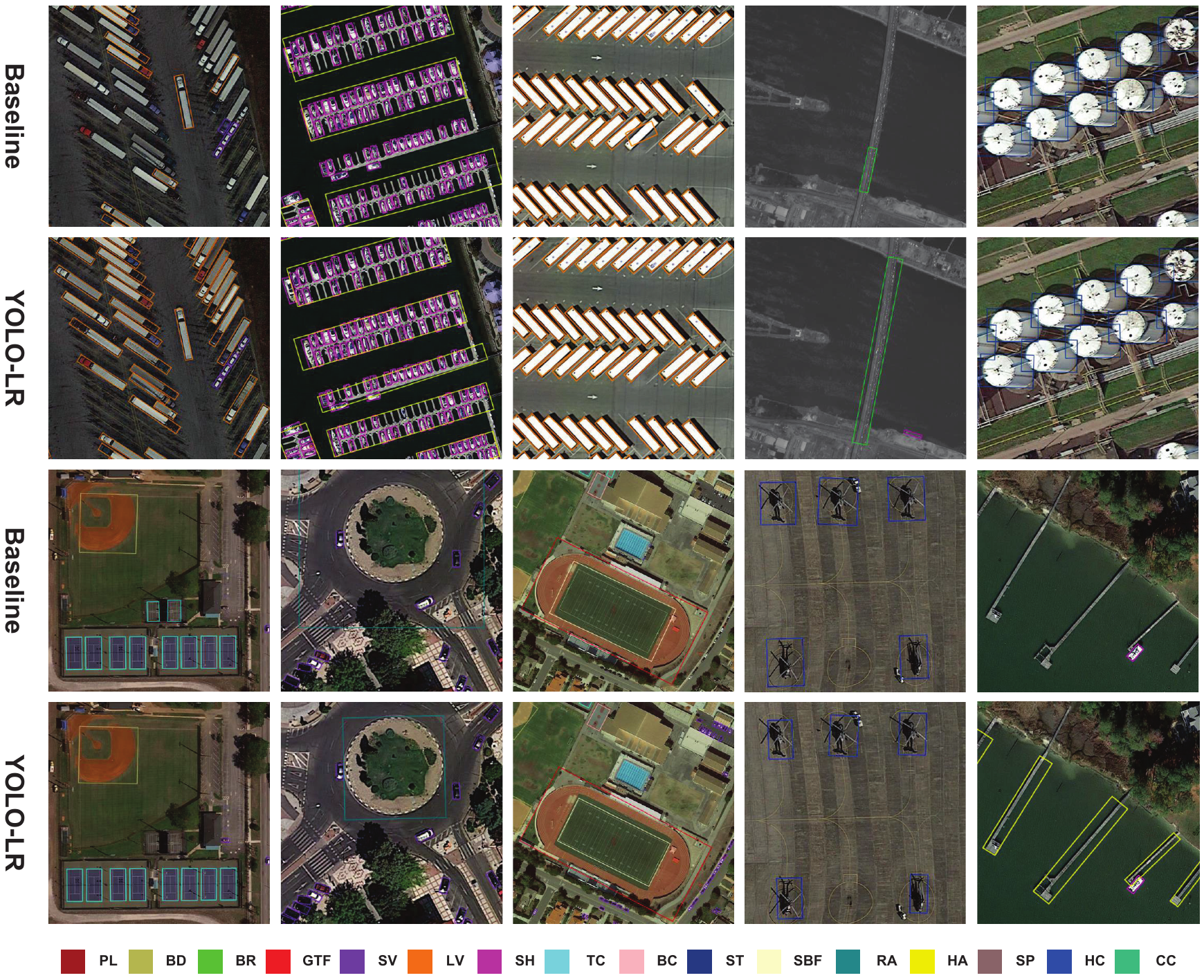}
  \caption{Some detection results of the baseline model and the proposed YOLO-LR on the DOTA-v1.5 dataset. The prediction results of YOLO-LR have fewer missed and false detection objects.}
  \label{fig:DOTA_display}
\end{figure*}

\subsubsection{Oriented Object Detection on the DOTA-v1.5 Dataset} 
We further apply the proposed LGBB and RRC to existing object detectors to evaluate their effectiveness. The regression target of original object detector is replaced by LGBB. RRC is applied to the first layer of each downsampling stage from $P_2$ to $P_6$. Multi-scale training and testing are conducted on the DOTA-v1.5 dataset. Some models are retrained based on the YOLOv7 and mmrotate framework.

As shown in Table \ref{table:DOTA_detection}, LGBB and RRC effectively enhance the detection performance of existing oriented object detectors in terms of the OBB representation and rotation-sensitive feature extraction. By applying the proposed LGBB and RRC, Oriented RCNN achieves the highest detection accuracy of 77.51\% mAP. Compared with the original method, the detection accuracy of the improved Oriented RCNN is increased by 0.48\% mAP. As designed in \cite{S2ANet}, we choose RetinaNet-O \cite{focal_loss} as the baseline model. Comparison detection results of the proposed detector YOLO with  LGBB and RRC (YOLO-LR) and the baseline model are visualized in Fig. \ref{fig:DOTA_display}. 

\begin{table}[t]
  \footnotesize
  \begin{center}
  \begin{tabular}{m{12.0em}<{\centering} m{5em}<{\centering} m{4.5em}<{\centering} m{4.5em}<{\centering}}
    \hline
    Method & Backbone & mAP(07) & mAP(12) \\ 
    \hline
    \multirow{1}{*}{R2CNN\cite{R2CNN}} & R101-FPN & 73.07 & 79.73\\
    \multirow{1}{*}{RoI Transormer\cite{RoI_Transformer}} & R101-FPN & 86.20 & -\\
    \multirow{1}{*}{Gliding Vertex\cite{Gliding_Vertex}} & R101-FPN & 88.20 & -\\
    \multirow{1}{*}{CenterMap-Net\cite{CenterMap}} & R50-FPN & - & 92.80\\
    \multirow{1}{*}{RetinaNet-O\cite{focal_loss}} & R101-FPN & 89.18 & 95.21\\
    \multirow{1}{*}{PIOU\cite{PIOU}} & DLA-34 & 89.20 & -\\
    \multirow{1}{*}{R3Det\cite{R3Det}} & R101-FPN & 89.26 & 96.01\\
    \multirow{1}{*}{R3Det-DCL\cite{DCL}} & R101-FPN & 89.46 & 96.41\\
    \multirow{1}{*}{CSL\cite{CSL}} & R101-FPN & 89.62 & 96.10\\
    \multirow{1}{*}{S2A-Net\cite{S2ANet}} & R101-FPN & 90.17 & 95.01\\
    \hline

    \multirow{1}{*}{YOLO \cite{YOLOv7}-LR} & YOLO & \textbf{90.19} & \textbf{95.55}\\
    \hline

    \multirow{1}{*}{Oriented Reppoints\cite{Oriented_reppoints}} & R50-FPN & 90.38 & 97.26\\
    \multirow{1}{*}{Oriented Reppoints\cite{Oriented_reppoints}-R} & R50-FPN & \textbf{90.49} & \textcolor{blue}{\textbf{97.68}}\\
    \hline

    \multirow{1}{*}{Oriented RCNN\cite{Oriented_RCNN}} & R50-FPN & 90.40 & 96.50\\
    \multirow{1}{*}{Oriented RCNN\cite{Oriented_RCNN}-LR} & R50-FPN & \textcolor{blue}{\textbf{90.52}} & \textbf{97.14}\\
    \hline

  \end{tabular}
  \end{center}

  \caption{Comparison results on the HRSC2016 dataset. mAP(07) and mAP(12) represent the results under VOC2007 and VOC2012 metrics, respectively.}
  \label{table:HRSC_detection}
\end{table}

\begin{table}[t]
  \footnotesize
  \begin{center}
  \begin{tabular}{m{12.0em}<{\centering} m{5em}<{\centering} m{4.5em}<{\centering} m{4.5em}<{\centering}}
    \hline
    Method & Backbone & FPS & mAP(\%) \\ 
    \hline
    \multirow{1}{*}{RetinaNet-O\cite{focal_loss}} & R101-FPN & 23.5 & 66.30\\
    \multirow{1}{*}{Faster RCNN-O\cite{DOTA}} & R101-FPN & 21.7 & 67.72\\
    \multirow{1}{*}{Gliding Vertex\cite{Gliding_Vertex}} & R101-FPN & 21.3 & 71.65\\
    \multirow{1}{*}{CSL\cite{CSL}} & R101-FPN & 22.1 & 72.55\\
    \multirow{1}{*}{R3Det\cite{R3Det}} & R101-FPN & 17.8 & 73.49\\
    \multirow{1}{*}{RoI Transformer\cite{RoI_Transformer}} & R101-FPN & 18.0 & 75.71\\
    \multirow{1}{*}{S2A-Net\cite{S2ANet}} & R101-FPN & 21.0 & 75.94\\
    \hline

    \multirow{1}{*}{YOLO \cite{YOLOv7}-LR} & YOLO & \textcolor{blue}{\textbf{24.3}} & \textbf{76.57}\\
    \hline

    \multirow{1}{*}{Oriented Reppoints\cite{Oriented_reppoints}} & R101-FPN & 17.6 & 76.59\\
    \multirow{1}{*}{Oriented Reppoints\cite{Oriented_reppoints}-R} & R101-FPN & \textbf{12.9} & \textbf{76.84}\\
    \hline

    \multirow{1}{*}{Oriented RCNN\cite{Oriented_RCNN}} & R101-FPN & 18.3 & 77.03\\
    \multirow{1}{*}{Oriented RCNN\cite{Oriented_RCNN}-LR} & R101-FPN & \textbf{11.8} & \textcolor{blue}{\textbf{77.51}}\\
    \hline

  \end{tabular}
  \end{center}

  \caption{Comparison results of speed and accuracy on the DOTA-v1.5 dataset.}
  \label{table:speed_vs_accuracy}
\end{table}

\subsubsection{Oriented Object Detection on the HRSC2016 Dataset}
For the HRSC2016 dataset, we replace the relative parts of some detectors with LGBB and RRC in Table \ref{table:HRSC_detection}. Some models are retrained based on the YOLOv7 and mmrotate framework. By integrating the proposed LGBB and RRC into the model structure, Oriented RCNN achieves the highest detection accuracy, reaching 90.52\% mAP under VOC2007 metric. By adding RRC, Oriented Reppoints achieves 97.68\% mAP under VOC2012 metric, which is 0.42\% mAP higher than the original method. Similar to the experimental results on the DOTA-v1.5 dataset, experiments on the HRSC2016 dataset demonstrate that LGBB and RRC are effective for oriented object detection. 

\subsection{Speed versus Accuracy}
The relationship between detection speed and detection accuracy on the DOTA-v1.5 dataset is presented in Table \ref{table:speed_vs_accuracy}. The model inference time includes Non-Maximum Suppression (NMS) time. A single RTX 3080Ti is deployed for inference with the input image size of $1024 \times 1024$. 

Compared with other methods, YOLO-LR achieves the highest detection speed (24.3 FPS) and competitive accuracy (76.57\% mAP). Applying LGBB and RRC to Oriented Reppoints and Oriented RCNN effectively improves model detection accuracy, and the loss of inference speed for these models is acceptable. The experimental results indicate that LGBB and RRC effectively improve model detection accuracy while maintaining a reasonable trade-off with detection speed. Therefore, LGBB and RRC can be extended to various oriented detection detectors.

\section{Conclusions}
\label{Conclusions}
This paper proposes a novel OBB representation, i.e., linear Gaussian bounding box (LGBB), and ring-shaped rotated convolution (RRC) for oriented object detection. By linearly transforming the elements of GBB, LGBB does not have the boundary discontinuity problem of OBB representations and has high numerical stability. By adaptively rotating feature maps to arbitrary orientations, RRC extracts rotation-sensitive features under a ring-shaped receptive field, rapidly aggregating rotation-sensitive features and contextual information. RRC can be applied to various models in a plug-and-play manner. Experimental results verify that the proposed LGBB and RRC achieve SOTA performance. By applying LGBB and RRC, various models achieve higher detection accuracy and have a reasonable trade-off for detection speed. In the future, we plan to develop a visual foundation model for rotation-sensitive feature extraction.


\bibliographystyle{elsarticle-num}
\bibliography{PR.bib}

\end{document}